\documentclass[accepted]{uai2022} 

\usepackage[american]{babel}

\usepackage{natbib} 
\usepackage{mathtools} 
\usepackage{booktabs} 
\usepackage{tikz} 

\usepackage[utf8]{inputenc}
\usepackage{graphicx}
\usepackage{textcomp}
\usepackage{xcolor}
\usepackage{multirow}
\usepackage{amsfonts}
\usepackage{array}
\usepackage{tabularx}

\title{Variational Autoencoders Without the Variation}

\author[1]{\href{mailto:<gd351@exeter.ac.uk>?Subject=Your UAI 2022 paper}{Gregory~A.~Daly}{}}
\author[1]{Jonathan~E.~Fieldsend}
\author[1]{Gavin~Tabor}

\affil[1]{%
    College of Engineering, Mathematics and Physical Sciences \\
    Harrison Building, Streatham Campus\\
    University of Exeter \\
    North Park Road \\
    Exeter, UK, EX4 4QF
}

\begin{document}
\maketitle

\begin{abstract}

Variational autoencdoers (VAE) are a popular approach to generative modelling. However, exploiting the capabilities of VAEs in practice can be difficult. Recent work on regularised and entropic autoencoders have begun to explore the potential, for generative modelling, of removing the variational approach and returning to the classic deterministic autoencoder (DAE) with additional novel regularisation methods. In this paper we empirically explore the capability of DAEs for image generation without additional novel methods and the effect of the implicit regularisation and smoothness of large networks. We find that DAEs can be used successfully for image generation without additional loss terms, and that many of the useful properties of VAEs can arise implicitly from sufficiently large convolutional encoders and decoders when trained on CIFAR-10 and CelebA.

\end{abstract}

\section{Introduction}\label{sec:intro}

Autoencoders are one of the early forms of neural networks, consisting of an encoder and decoder function, \(z=f(x)\) and \(r=g(z)\), where there is a some kind of bottleneck that forces the model to learn a useful latent representation in \(z\). The encoder and decoder can take any form, such as deep neural networks, convolutional neural networks (CNN) or recurrent neural networks. Early autoencoders were used for dimensionality reduction, feature extraction and learning manifolds. Stochastic approaches were also developed to improve the ability of early autoencoders to improve their ability to create new samples by generative approaches \citep{goodfellowDeepLearning2016}. Early autoencoders were difficult to use for generative modelling as the distribution of the information in the latent space could not be described with a simple prior function and the latent spaces were not particularly smooth or interpolatable. In short, it was not clear what posterior distribution to sample from to generate new data, and the space between trained points was not meaningful enough.

Variational Bayesian approaches give a method to approximate the intractable posterior and \citet{kingmaAutoEncodingVariationalBayes2014} developed a method to use this approach to create the variational autoencoder (VAE). VAEs have become one of the major areas of research in generative modelling and have supplanted other autoencoder approaches. VAEs have been used to create state-of-the-art generative models \citep{vahdatNVAEDeepHierarchical2021}, to produce more efficient and performative reinforcement learning agents \citep{nairVisualReinforcementLearning2018} and drug discovery \citep{jinJunctionTreeVariational2018}.

The VAE approach of \citet{kingmaAutoEncodingVariationalBayes2014} and \citet{rezendeStochasticBackpropagationApproximate2014} allows us to train a generative model \(p(\mathbf{z|x}) = p(\mathbf{z})p(\mathbf{x|z})/p(\mathbf{x})\) where \(\mathbf{x}\) is our input data, \(\mathbf{z}\) is the projection of those data in the latent space, \(p(\mathbf{z})\) is a prior distribution over the latent variables and \(p(\mathbf{x|z})\) is a decoder to generate data from latent variables. The true posterior, \(p(\mathbf{z|x})\) is intractable, and so is approximated by a probabilistic encoder \(q(\mathbf{z|x})\).  The prior is assumed to be a centred isotropic multivariate Gaussian \(p(\mathbf{z}) = \mathcal{N} (\mathbf{z; 0, I})\) with a diagonal covariance matrix.

On its own, this approach does not have a tractable derivative and cannot be trained by gradient descent. The encoder and decoder are jointly trained by using the reparameterisation trick to make \(z\) a deterministic variable with some additional external noise, \(\epsilon\). The model can then be trained by gradient descent to maximise the evidence lower bound:
\begin{equation}\label{eq:elbo}\nonumber
\mathbf{L}(q) = \mathbb{E}_{z\sim q(\mathbf{z|x})}\log p(\mathbf{x|z}) - \beta D_{KL}(q(\mathbf{z|x})||p(\mathbf{z})).
\end{equation}

These two parts of the loss act as a regularisation and reconstruction error term respectively and \(\beta\) is a weight term to balance the strength of the two parts. \(D_{KL}\) calculates the Kullback-Leibler divergence between the two distributions \(q(\mathbf{z|x})\) and \(p(\mathbf{z})\), which in practice acts as a strong regulariser to encourage the distribution of \(\mathbf{z}\) to follow the Gaussian prior.

Although VAEs have become a widely used framework for generative modelling, they have suffered from problems that have been a focus of much research. The priors of VAEs do not generally match their assumed Gaussian posteriors \citep{kingmaImprovedVariationalInference2016} and many alternative approaches have been postulated to improve on this \citep{kingmaImprovedVariationalInference2016,tomczakVAEVampPrior2018,daiDiagnosingEnhancingVAE2019}. Overly strong regularisation in the VAE loss can cause posterior collapse, where the values in the latent space collapse to a single value due to the overly strong regularisation of \(D_{KL}\) \citep{vandenoordNeuralDiscreteRepresentation2017, heLaggingInferenceNetworks2019}. The VAE objective can also learn latent spaces that have trivial solutions or have not autoencoded the data \citep{chenVariationalLossyAutoencoder2017, zhaoDeeperUnderstandingVariational2017}. Other work has focused on improving the performance of VAEs by finding optimal encoder and decoder architectures \citep{vahdatNVAEDeepHierarchical2021}, using new loss functions based on Wasserstein distance \citep{tolstikhinWassersteinAutoEncoders2018} and using discrete latent spaces \citep{vandenoordNeuralDiscreteRepresentation2017,razaviGeneratingDiverseHighFidelity2019}.

Some recent work has begun to look at the generative performance of autoencoders without the variational inference layer -- deterministic autoencoders (DAE). \citet{ghoshVariationalDeterministicAutoencoders2019} introduced a regularised autoencoder (RAE), training a classical AE with a reconstruction loss and an additional regularisation term. \citet{ghoseBatchNormEntropic2020} introduced an entropic autoencoder (EAE), applying batch normalisation to the latent space to constrain its mean and variance and using a new regularisation method. \citet{saseendranShapeYourSpace2021} introduced a new loss function that can guide the latent representation of DAEs towards a desired latent prior. 

There is an interesting result in these papers that has not been expanded upon, ordinary autoencoders, without additional regularisation or novel loss terms, achieved comparable performance to their proposed methods and beat VAEs and Wasserstein autoencoders (WAE), when the assumption of the latent prior being a centred isotropic multivariate Gaussian \(p(\mathbf{z}) = \mathcal{N} (\mathbf{z; 0, I})\) with a diagonal covariance matrix is dropped. We will investigate this interesting result empirically in this paper and try to place it in context with the wider research around the generalisability and interpolatability of neural networks.

\section{Related Work}

It is worth looking in more detail at the recent approaches to DAEs to understand the motivation of their proposed alterations to improve generative performance.  \citet{ghoshVariationalDeterministicAutoencoders2019} posit that a VAE can be formulated as a AE with Gaussian noise added to the decoder input and then propose that this can be replaced with regularisation applied to the decoder. Earlier work from \citet{alainWhatRegularizedAutoencoders2014, bengioRepresentationLearningReview2014, rifaiGenerativeProcessSampling2012} applied regularisation to contractive autoencdoers and also proposed methods to generate samples from them. However, \citet{ghoshVariationalDeterministicAutoencoders2019} argues that these networks were hard to train and tune and their method has similar properties while being much more computational efficient. By removing the variational inference layer they lose the simple ability to sample their prior from \(p(\mathbf{z}) = \mathcal{N} (\mathbf{z; 0, I})\) and propose to fit a ex-post density estimator over the latent space. The two used are a 10 component Gaussian mixture model (GMM) and a full covariance multivariate Gaussian, dropping the VAE assumptions of a diagonal covariance matrix and a \(\mu = 0 , \sigma = 1\). This is justified by previous work showing that VAE posteriors do not match these assumptions on the prior and improved performance can be achieved with different priors \citep{daiDiagnosingEnhancingVAE2019}. The surprisingly good performance of an unmodified AE is discussed in Appendix I of \citet{ghoshVariationalDeterministicAutoencoders2019}, where it is suggested that convolutional neural networks and gradient-based optimisers provide some inherent regularisation.

\citet{ghoseBatchNormEntropic2020} extends from \citet{ghoshVariationalDeterministicAutoencoders2019} by applying batch normalisation (BN) to the latent representation layer after the encoder and using an entropic regularisation term in the loss function, in a similar fashion to the Jacobian regularisation term used by \citet{rifaiGenerativeProcessSampling2012}. The motivation of this is to regain the ability to sample \(p(\mathbf{z}) = \mathcal{N} (\mathbf{z; 0, I})\) without the need for the ex-post density estimator of RAEs. This work outperformed RAEs when using the same ex-post density estimation techniques and outperformed VAEs and WAEs when sampling from \(\mathcal{N} (\mathbf{z; 0, I})\). In Section 4.2 of their paper \citet{ghoseBatchNormEntropic2020} makes an important prediction, that in a deep, narrow-bottlenecked AE, with BN applied to the latent layer, the latent space's true prior will approach a spherical Gaussian.  \citet{ghoseBatchNormEntropic2020} also noted that classic AEs are sensitive to the number of units in the latent space and having a narrower bottleneck will ``\emph{incentivize Gaussianization}'' by encouraging each latent dimension to carry more information, and this is viewed as supporting their prediction in Section 4.2 of their paper.

\citet{saseendranShapeYourSpace2021} introduce an extension to \citet{ghoshVariationalDeterministicAutoencoders2019} by developing a novel loss function that removes the need for the ex-post density estimation step. This is achieved by a new loss function that pushes the latent representation towards some pre-determined prior, such as a GMM. The performance is similar to \citet{ghoshVariationalDeterministicAutoencoders2019}, less than \citet{ghoseBatchNormEntropic2020} and similar to ordinary AEs. \citet{saseendranShapeYourSpace2021} does not investigate or comment on the performance of ordinary AEs in comparison to their method.

\subsection{Wider related work}\label{sec:widerwork}

Batch normalisation (BN) was introduced by \citet{ioffeBatchNormalizationAccelerating2015} as a method to reduce internal covariate shift in deep networks and demonstrated significant improvements to the accuracy of inception networks. However, other work in the area have empirically shown that BN can increase covariate shift and have offered alternative explanations for BN's improvements of network accuracy, such as improving the smoothness of the loss landscape \citep{santurkarHowDoesBatch2018}. Alternative normalisation techniques have also been proposed in GAN research, \citet{miyatoSpectralNormalizationGenerative2018} proposed spectral normalisation (SN) as an alternative technique that by normalising the weight matrix to satisfy a Lipschitz constraint to achieve better generalised performance on GAN image generation. SN was applied to convolutional layers in both RAEs and EAEs. \citet{baLayerNormalization2016} introduced layer normalisation (LN) as an alternative to BN that is not sensitive to the batch size during training and applies the same calculation during training and evaluation.

Interpolatable manifolds with enough smoothness are a key property of any prior and latent space for generative modelling. There is a growing body of research on understanding the origins and behaviour of smoothness in neural networks, and we will review a small part of that literature to provide some context on how this can affect all types of autoencoder.

\citet{barrettImplicitGradientRegularization2020} provide an analysis of the implicit regularisation affect of gradient-based optimisation. They are able to demonstrate that gradient descent has an implicit regularisation term that is proportional to the learning rate and network size. This is an important observation in the context of VAEs as the KL-divergence term provides strong regularisation to encourage the posterior to fit the assumed prior. As the posterior does not generally match the Gaussian prior \citep{kingmaImprovedVariationalInference2016}, this regularisation affect in small networks may not be inherently necessary in a larger network where the implicit regularisation from gradient based optimisation is strong enough.


\citet{bubeckUniversalLawRobustness2021} investigate how the size and dimensionality of the models and data affect the ability of a model to produce a smooth interpolation of the learned data. They propose a lower bound on the on the number of parameters required to produce a smoothly interpolatable model that is proportional to the number of parameters in the model and the dimensionality of the data set. This is an important finding in the context of VAEs as it suggests that the encoder will be able to produce a latent space with better smoothness and interpolatability and the decoder will be better able to map similar latent codes back into similar areas of the input data space with increasing size and decreasing dimensionality of the data. 

Another consequence of this, as discussed in Section 1.3 of \citet{bubeckUniversalLawRobustness2021}, is that, as a consequence of isoperimetry, high dimensional Lipschitz continuous functions are concentrated around their means and a high dimensional multivariate Gaussian can satisfy this. This would suggest that approaching this lower bound could also results in networks that produce outputs that are closer to a smooth Gaussian.

\subsection{Our contribution}\label{sec:hypotheses}

Our hypotheses are:
\begin{enumerate}
    \item Variational inference is required to create smooth and interpolatable latent spaces with weak/shallow encoders and decoders through strong regularisation. Powerful/deep encoders and decoders intrinsically produce smooth and interpolatable latent spaces and this will dominate the behaviour of the autoencoder.\label{itm:1sthyp}
    \item Powerful/deep encoders produce outputs that tend towards smooth Gaussians as they increase in depth and complexity\label{itm:2ndhyp}
    \item For a given power of encoder and decoder and complexity/dimensionality of the training data, there is an optimum size of the bottleneck layer that balances quality of reconstruction and generative samples, implicitly balancing the regularising capability of deep networks and the ability of the autoencoder to overfit the data.\label{itm:3rdhyp}
\end{enumerate}

We will investigate this empirically by looking at the encoder-decoder architecture used in \citet{tolstikhinWassersteinAutoEncoders2018, ghoshVariationalDeterministicAutoencoders2019, ghoseBatchNormEntropic2020} restricting ourselves to using only reconstruction loss for gradient based optimisation. We will show the effect of encoder-decoder depth and width, bottleneck size and other augmentations such as normalisation layers.

\section{Experiments}


\begin{table}[t!]
\caption{List of experiments.}
\label{tab:experimentlist}
\centering
\begin{tabular}{m{0.75cm}m{1.05cm}m{1.4cm}m{0.8cm}m{2.1cm}}
\toprule
Filters & Depth & Dataset & Model & Other \\ 
\midrule
32 & 1 & CIFAR-10, CelebA & DAE, VAE &   \\ 
\midrule
\multirow{2}{*}{64}  & 1 & CIFAR-10, CelebA & DAE, VAE &   \\
 & 2, 3 & CIFAR-10 & DAE &  \\ 
\midrule
\multirow{2}{*}{128} & 1 & CIFAR-10, CelebA & DAE, VAE & BN, SN, LN, Gaussian noise, trainable latent \\
 & 2, [3,3,2,1] & CIFAR-10 & DAE & \\ 
\midrule
256 & 1 & CIFAR-10 & DAE, VAE & \\
\bottomrule

\end{tabular}
\end{table}

\begin{figure*}[h]
    \centering
    \includegraphics[width=\textwidth, page = 4]{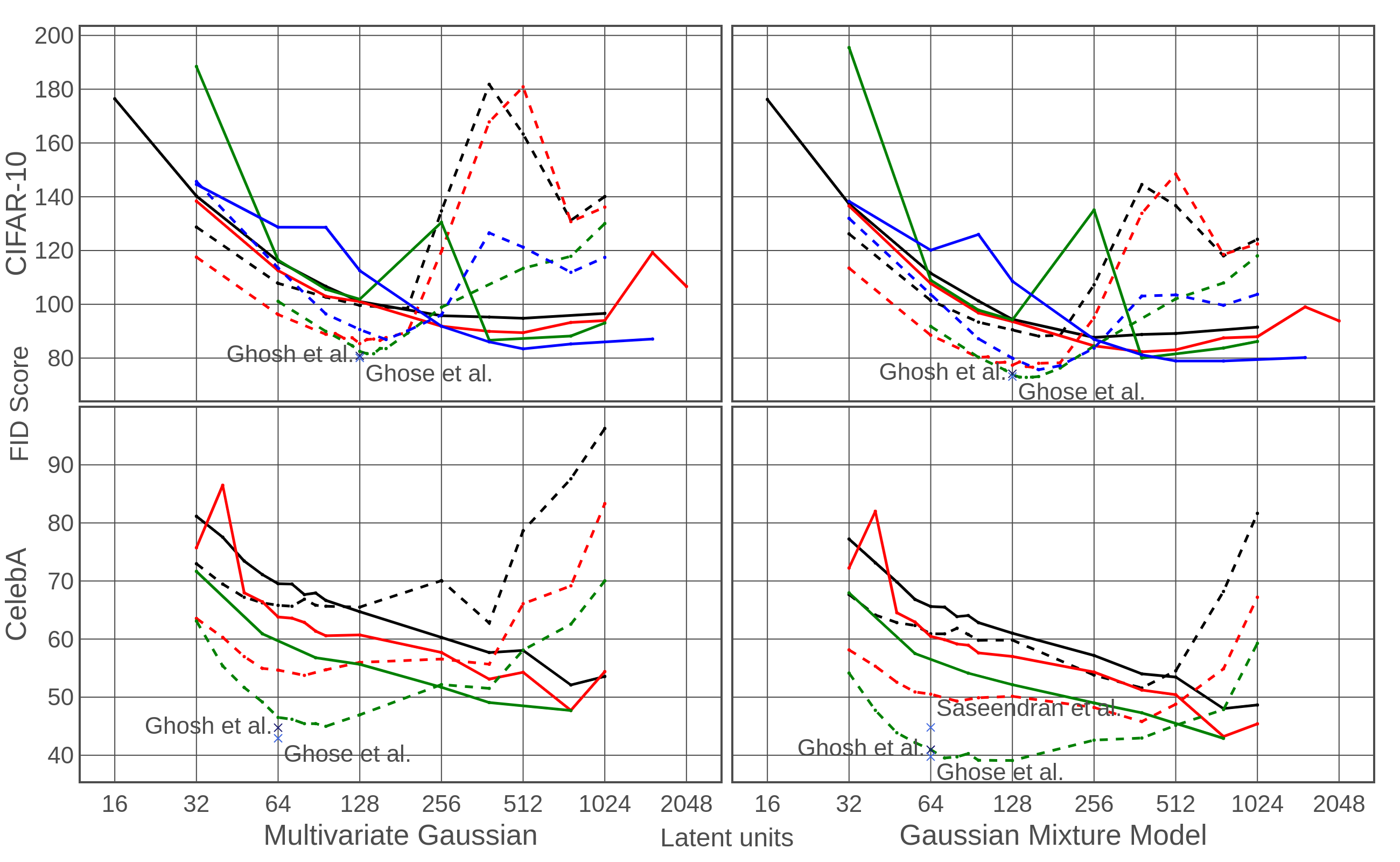}
    \caption{FID Score on CIFAR-10 and CelebA for DAE and VAEs with varying size of the latent space. VAEs are solid lines, DAEs are dotted, filter bases are (32, black), (64, red), (128, green), (256, blue). The left graph shows FIDs for samples drawn from a MVG and a GMM on the right. Additional crosses show the comparative best results from \citet{ghoshVariationalDeterministicAutoencoders2019, ghoseBatchNormEntropic2020, saseendranShapeYourSpace2021}.}
    \label{fig:DAEvsVAE}
\end{figure*}

The majority of our experiments will be carried out with the CIFAR-10 data set \citep{krizhevskyLearningMultipleLayers2009}, as is the case with other works \citep{daiDiagnosingEnhancingVAE2019, vahdatNVAEDeepHierarchical2021, tolstikhinWassersteinAutoEncoders2018, ghoshVariationalDeterministicAutoencoders2019, ghoseBatchNormEntropic2020}. CIFAR-10's small size allows us to explore a wider range of network configurations within a reasonable computational envelope, whilst a subset of experiments will be repeated with the CelebA data set \citep{liuDeepLearningFace2015}. We will use the same basic convolutional architecture used in \citet{tolstikhinWassersteinAutoEncoders2018, ghoshVariationalDeterministicAutoencoders2019, ghoseBatchNormEntropic2020} and vary the width, depth, and types of normalisation layers, as well as the addition of Gaussian noise and freezing layers. A subset of these experiments will be repeated with a VAE latent layer and objective function. The convolutional encoder consists of four layers doubling the number of filters each time and the decoder consists of two layers of 4 times and 2 times the initial number of filters. Strided convolutions are used to upscale and downscale layers when the filter number changes and 4x4 filter kernels will be used for CIFAR-10 and 5x5 for CelebA. For larger depths, the convolutional layers are repeated \(n\) times, with the strided convolution last.

\begin{table*}[ht]
\centering
\caption{FID Scores for VAE and DAE architectures for 64 and 128 latent units on CelebA and CIFAR-10 respectively. Train R and Test R are the FID scores of the reconstructed samples. Bold values are the lowest score for each filter width.}
\label{tab:scoresat128}
\begin{tabular}{m{0.8cm}m{0.9cm}|m{1.1cm}m{1cm}m{1.3cm}m{0.8cm}m{0.8cm}|m{1.1cm}m{1cm}m{1.3cm}m{0.8cm}m{0.8cm}}
\toprule
\multirow{2}{*}{Filters} & \multirow{2}{1.2cm}{Model} & \multicolumn{5}{c}{CIFAR-10} & \multicolumn{5}{c}{CelebA} \\ 
 & & Train R & Test R & \(\mathcal{N} (\mathbf{z; 0, 1})\) & MVG & GMM & Train R & Test R & \(\mathcal{N} (\mathbf{z; 0, 1})\) & MVG & GMM \\ 
\midrule
32 & DAE & \textbf{50.56}   & \textbf{60.20}  & 288.8 & \textbf{99.60}  & \textbf{90.55}  & \textbf{49.48}   & \textbf{50.38}  & 346.4 & 65.50 & \textbf{59.83} \\
32 & VAE & 62.46   & 79.06  & \textbf{100.1} & 101.0 & 94.40  & 51.95   & 53.21  & \textbf{64.53}  & \textbf{64.71} & 61.01 \\
\midrule
64 & DAE  & \textbf{30.94}   & \textbf{48.05}  & 372.30 & \textbf{85.38}  & \textbf{77.36}  & \textbf{41.78}   & \textbf{43.68}  & 370.9 & \textbf{55.99} & \textbf{50.15} \\
64 & VAE & 56.10   & 78.35  & \textbf{99.49}  & 101.0 & 93.60  & 48.65   & 50.86  & \textbf{60.63}  & 60.71 & 57.01 \\
\midrule
128 & DAE & \textbf{16.95}   & 45.21  & 282.3 & 82.49  & 73.78  & \textbf{29.01}   & \textbf{31.85}  & 426.9 & 46.96 & \textbf{39.10} \\
128 & DAE (BN) & 17.69 & 45.24  & 94.14  & 82.64  & \textbf{73.05} & -- & -- & -- & -- & -- \\
128 & VAE & 48.41   & 76.20  & 99.62  & 101.9 & 94.21  & 43.38   & 46.44  & 55.70  & 55.66 & 52.15 \\
128 & WAE & -- & 57.94  & 117.4 & -- & 93.53  & -- & 34.81  & 53.67  & -- & 42.73 \\
128 & RAE & -- & 32.24  & -- & \textbf{80.80}  & 74.16  & -- & 36.01  & -- & \textbf{44.74} & 40.95 \\
128 & EAE & -- & \textbf{29.77}  & \textbf{85.26}  & -- & 73.12  & -- & 40.26  & \textbf{44.63}  & -- & 39.76 \\
128 & GMM - DAE & -- & -- & -- & -- & -- & -- & 39.48  & -- & 49.79 & 44.79 \\
\midrule
256 & DAE & \textbf{8.07}    & \textbf{47.51}  & 402.7 & \textbf{90.59}  & \textbf{79.97}  & -- & -- & -- & -- & -- \\
256 & VAE & 50.85   & 93.49  & \textbf{112.0} & 112.5 & 108.6 & -- & -- & -- & -- & -- \\
\bottomrule
\end{tabular}
\end{table*}

CelebA images are centre cropped to 140x140 and downscaled to 64x64. All models are trained on Tensorflow v.2.4 with a single NVIDIA RTX2080Ti GPU and with Tensorflow, Numpy, and Python random seeds fixed. Training was for 100 epochs with a batch size of 128 using Adam with a piecewise learning rate decay schedule of \([10^{-3}, 10^{-4}, 10^{-5}]\) decaying at the 33rd and 66th epoch with one warmup epoch at a learning rate of \(10^{-4}\). DAEs used binary cross-entropy as the loss function and VAEs used binary cross-entropy with a \(\beta\)-weighted KL-divergence, where \(\beta=0.01\).

To evaluate the generative performance of all models the generative samples were evaluated by calculating FID scores with the Tensorflow Inception v3 model \citep{heuselGANsTrainedTwo2017}. Samples are generated by three methods, randomly sampling from a univariate Gaussian, \(\mathcal{N} (\mathbf{z; 0, 1})\), a multivariate Gaussian (MVG) using the means and full covariance matrix of the encoded training samples and a 10 component Gaussian mixture model (GMM) fit on the encoded training samples. To calculate the FID scores, 10000 samples were used with the 10000 CIFAR-10 test images or with 10000 randomly drawn test images for CelebA.

A list of all experiments is given in Table \ref{tab:experimentlist}. For DAEs and VAEs we have varied the number of convolutional filters in the network and trained on CelebA and CIFAR-10. For DAEs we have also varied the network depth for 64 and 128 filters, applied BN, SN and LN to the latent layer, applied Gaussian noise to the input and frozen the latent layer at its random initialisation for 128 filters and depths of 1 and 2 during training.

\section{Results}

The first set of results we will look at is the comparison between VAEs and DAEs in CIFAR-10 and CelebA with a depth of 1. In Figure \ref{fig:DAEvsVAE} we can see the relative performance of VAEs and DAEs for different numbers of filters sampling by MVG and GMM. In both cases DAEs always outperform VAEs at the comparison points of 128 and 64 latent units for CIFAR-10 and CelebA respectively. A summary of these results at the comparison point of 128 latent units can be seen in Table \ref{tab:scoresat128}. The best results from RAE and EAE approaches of \citet{ghoshVariationalDeterministicAutoencoders2019, ghoseBatchNormEntropic2020} are marginally beaten by or marginally beat un-altered DAEs on CIFAR-10 and CelebA with MVG and GMM sampling.

\begin{table}[ht!]
\centering
\caption{FID Scores for VAE and DAE architectures at best number latent units on CelebA and CIFAR-10 respectively. Test R is the FID scores of the reconstructed samples. Bold values are the lowest score for each filter width.}
\label{tab:bestscores}
\begin{tabular}{p{0.8cm}p{0.75cm}p{0.75cm}p{0.7cm}p{0.7cm}p{0.7cm}p{0.75cm}}
\toprule
\multicolumn{7}{c}{CIFAR-10} \\
\midrule
Latent units & Filters & Model   & Test R & Norm   & MVG    & GMM   \\
\midrule
160          & 32      & DAE     & 51.06  & 376.1 & 98.90  & 88.17 \\
192          & 32      & DAE     & \textbf{43.55}  & 419.6 & 98.49  & 88.51 \\
256          & 32      & VAE     & 62.69  & 96.14  & 95.77  & \textbf{87.67} \\
384          & 32      & VAE     & 56.07  & \textbf{95.29}  & 95.25  & 88.77 \\
512          & 32      & VAE     & 52.08  & 96.59  & \textbf{94.79}  & 89.14 \\
\midrule
128          & 64      & DAE     & 48.05  & 372.3 & \textbf{85.38}  & 77.36 \\
152          & 64      & DAE     & \textbf{43.22}  & 370.8 & 86.55  & \textbf{76.59} \\
128          & 64      & VAE     & 78.35  & 99.49  & 101.0 & 93.60 \\
384          & 64      & VAE     & 51.59  & \textbf{89.77}  & 89.92  & 82.35 \\
512          & 64      & VAE     & 46.01  & 90.95  & 89.45  & 83.06 \\
\midrule
144          & 128     & DAE     & \textbf{40.31}  & 428.6 & \textbf{81.62}  & \textbf{72.82} \\
384          & 128     & VAE     & 46.98  & \textbf{87.29}  & 86.65  & 80.01 \\
\midrule
160          & 256     & DAE     & 37.70  & 379.5 & 86.90  & \textbf{75.67} \\
512          & 256     & VAE     & 39.50  & \textbf{82.34}  & \textbf{83.45}  & 78.95 \\
768          & 256     & VAE     & \textbf{37.27}  & 87.68  & 85.25  & 78.91 \\
\midrule
\multicolumn{7}{c}{CelebA}                                          \\
\midrule
384          & 32      & DAE     & \textbf{32.50}  & 474.4 & 62.77  & 51.58 \\
768          & 32      & VAE     & 39.83  & \textbf{54.53}  & \textbf{52.10}  & \textbf{48.03} \\
\midrule
80           & 64      & DAE     & 45.83  & 409.5 & 53.76  & 49.31 \\
384          & 64      & DAE     & \textbf{29.62}  & 502.9 & 55.67  & 45.80 \\
768          & 64      & VAE     & 38.42  & \textbf{50.34}  & \textbf{47.73}  & \textbf{43.23} \\
\midrule
96           & 128     & DAE     & 34.84  & 406.47 & \textbf{44.98}  & \textbf{39.13} \\
128          & 128     & DAE     & \textbf{31.85}  & 426.9 & 46.96  & 39.10 \\
384          & 128     & VAE     & 40.04  & \textbf{49.09}  & 49.07  & 47.30 \\
768          & 128     & VAE     & 36.93  & 50.67  & 47.70  & 42.91 \\
\bottomrule
\end{tabular}
\end{table}

However, a fairer comparison is the relative performance of DAEs and VAEs at their optimum number of latent units; a summary of which is shown in Table \ref{tab:bestscores}. On CIFAR-10, the DAE outperforms the VAE for 64 and 128 filters and the VAE achieves better FIDs for 32 and 256 filters for MVG sampling and the DAE outperforms the VAE for 64, 128 and 256 filters with GMM sampling. On CelebA, VAEs are able to achieve better FIDs, with large numbers of latent units, for 32 and 64 filters but at 128 filters DAEs won out.  

Both VAEs and DAEs have a minimum in the number of latent units, but it was difficult to investigate the VAEs for more than 512 units as they began to become unstable during training, DAE training was simple and robust at all latent space sizes. The variation in DAE performance with the size of the latent space was much stronger, when using either VAEs or DAEs for generative or downstream tasks this is an important parameter to optimise, but even more so in DAEs. 

\begin{figure*}[h]
    \centering
    \includegraphics[width=\textwidth]{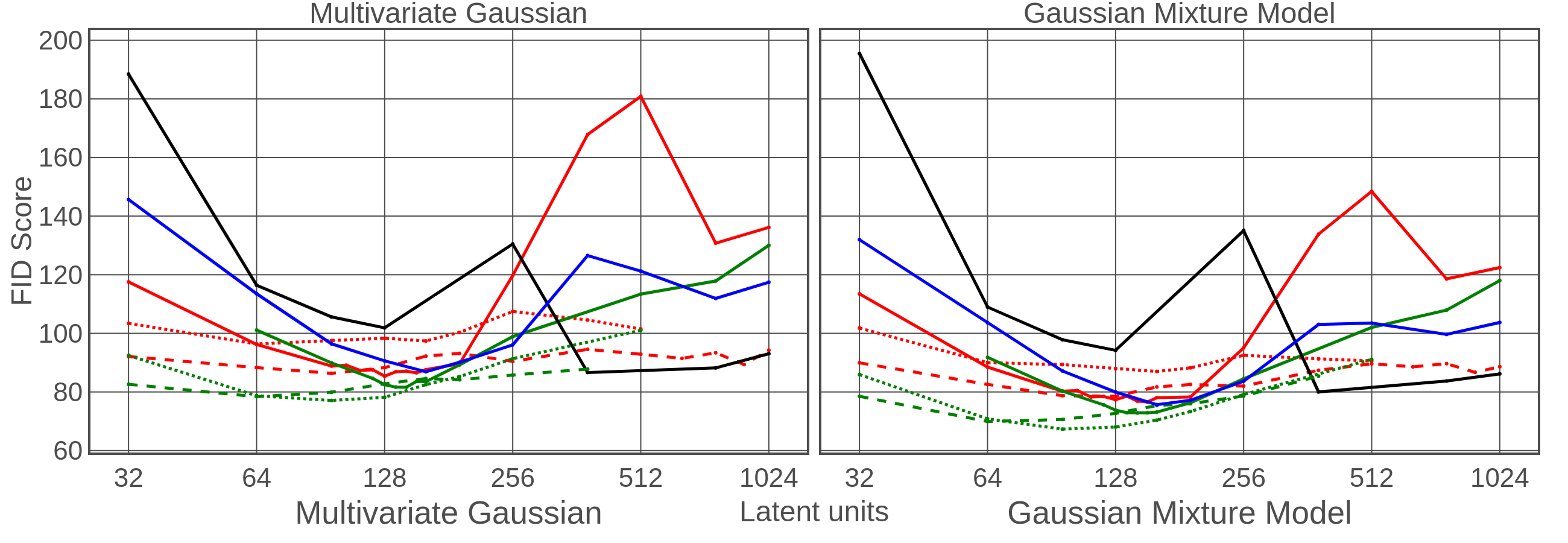}
    \caption{FID Score on CIFAR-10 for DAEs with varying size of the latent space and a VAE baseline. The solid black line is the VAE baseline, solid lines are a depth of 1, dash a depth of 2 and dotted lines are a depth of 3 for 64 filters and depths of [3,3,2,1] for 128 filters. Filters are (64, red), (128, green), (256, blue). The left graph shows FIDs for samples drawn from a MVG and a GMM on the right.}
    \label{fig:DAEfiltersanddepth}
\end{figure*}

\begin{figure}[h]
    \centering
    \includegraphics[width=0.49\textwidth]{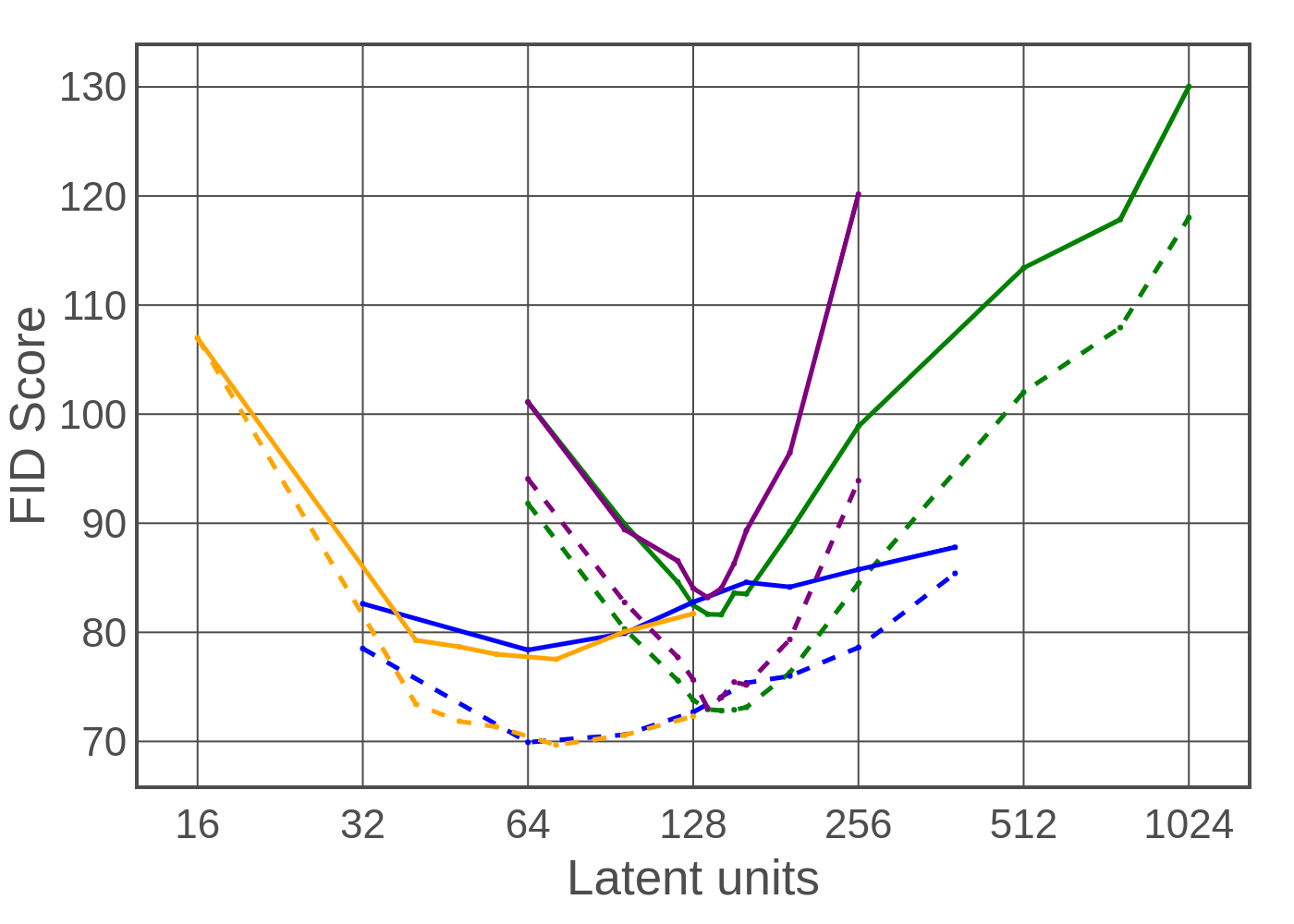} 
    \caption{FID Score on CIFAR-10 for DAEs with varying size of the latent space. Solid lines are FIDs for samples drawn from a MVG and dotted lines are for samples drawn from a GMM. Green and blue lines are DAEs with 128 filters and a depth of 1 and 2 respectively. Purple and orange lines are the same architecture, except the latent space is frozen at its random initialisation.}
    \label{fig:DAEfrozenlatents}
\end{figure}

One of the most interesting results is shown in Figure \ref{fig:DAEfrozenlatents}, a comparison between DAEs with 128 filters and a depth of 1 and 2 where the latent layer has been frozen at its random initialisation. Remarkably, given how sensitive the generative performance is to the size of the latent space, there is very little difference in performance when the latent layer is frozen in training.

In Figure \ref{fig:DAEfiltersanddepth}, when increasing the depth of the convolutional layers, the sensitivity of the generative performance to the number of latent units decreases significantly but does not uniformly lead to an increase in generative performance. For a filter base of 64 there is a drop in the best FID score, but for a filter base of 128 the deeper networks achieve better FID scores and the location of the minimum in the number of latent units significantly decreases. The best score achieved on CIFAR 10 was 77.10 for MVG and 67.32 for GMM on a DAE with 96 latent units, 128 filters and depths of [3,3,2,1] for the four layers.

In \citet{ghoshVariationalDeterministicAutoencoders2019}, the authors claim that a VAE can be interpreted as a DAE with noise injected at the decoder input. To test this we trained a DAE with the Tensorflow Gaussian Noise layer added to the decoder input and trained models with standard deviations of 0.1, 0.5 and 1 for the Gaussian Noise layer. The Gaussian Noise layer samples from a normal distribution of mean \(0\) and the standard deviation set for the layer and adds it to the input of the layers. Increasing noise levels degraded the performance of the DAE, both in terms of  reconstruction error and generative performance, and between standard deviations of 0.1 and 0.5 the DAE performance drops below that of the VAE baseline. This may provide some evidence for the claim in \citet{ghoshVariationalDeterministicAutoencoders2019}, but is not definitive.

\begin{figure*}[h]
    \centering
    \includegraphics[width=0.95\textwidth]{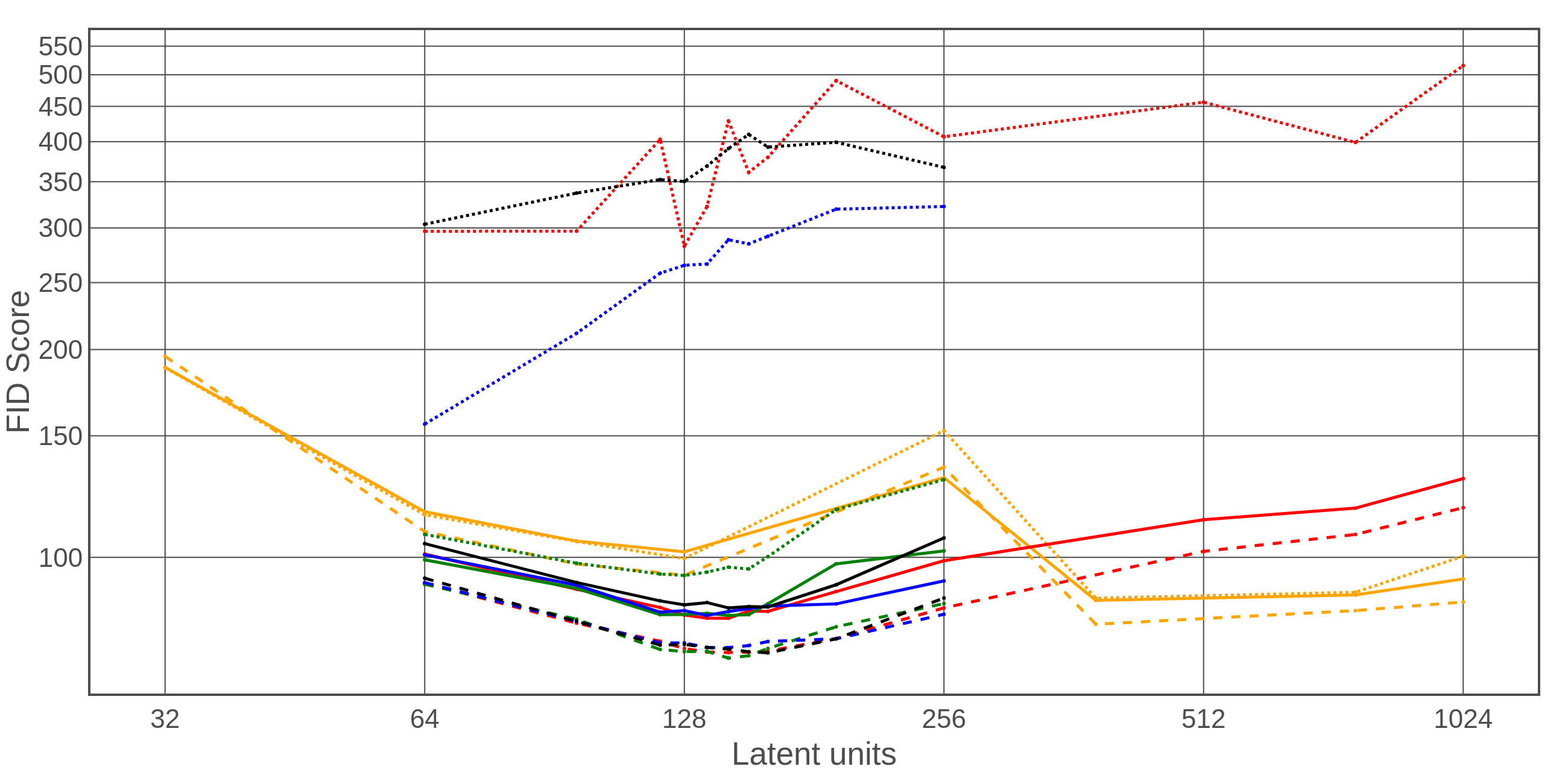}
    \caption{FID Score on CIFAR-10 for DAEs with depth 1 and 128 filters, varying the size of the latent space for different types of latent normalisation and a baseline VAE. Solid lines are MVG samples, dashed lines are GMM samples and dotted lines are sample from \(\mathcal{N} (\mathbf{z; 0, I})\). Red and orange lines are the DAE and VAE baselines respectively, green in Batch Normalisation, blue is Spectral Normalisation and black is Layer Normalisation.}
    \label{fig:DAElatentnorms}
\end{figure*}

In \citet{ghoseBatchNormEntropic2020}, a BN layer is applied at the end of the encoder and they train the model by minimising a new regularising objective function. In Section \ref{sec:widerwork} we discuss the current interpretations of BN and other layer normalisation approaches. When applying BN, SN and LN to the output of the encoder there was very little change in the performance of the DAE for MVG and GMM sampling, as can be seen in Figure \ref{fig:DAElatentnorms}. BN did have the affect of massively improving the generative performance when sampling from \(\mathcal{N} (\mathbf{z; 0, I})\), beating the baseline VAE, but not the EAE. This is in agreement with the results from \citet{ghoseBatchNormEntropic2020}. 

\begin{figure*}[t!]
    \centering
    \includegraphics[width=\textwidth]{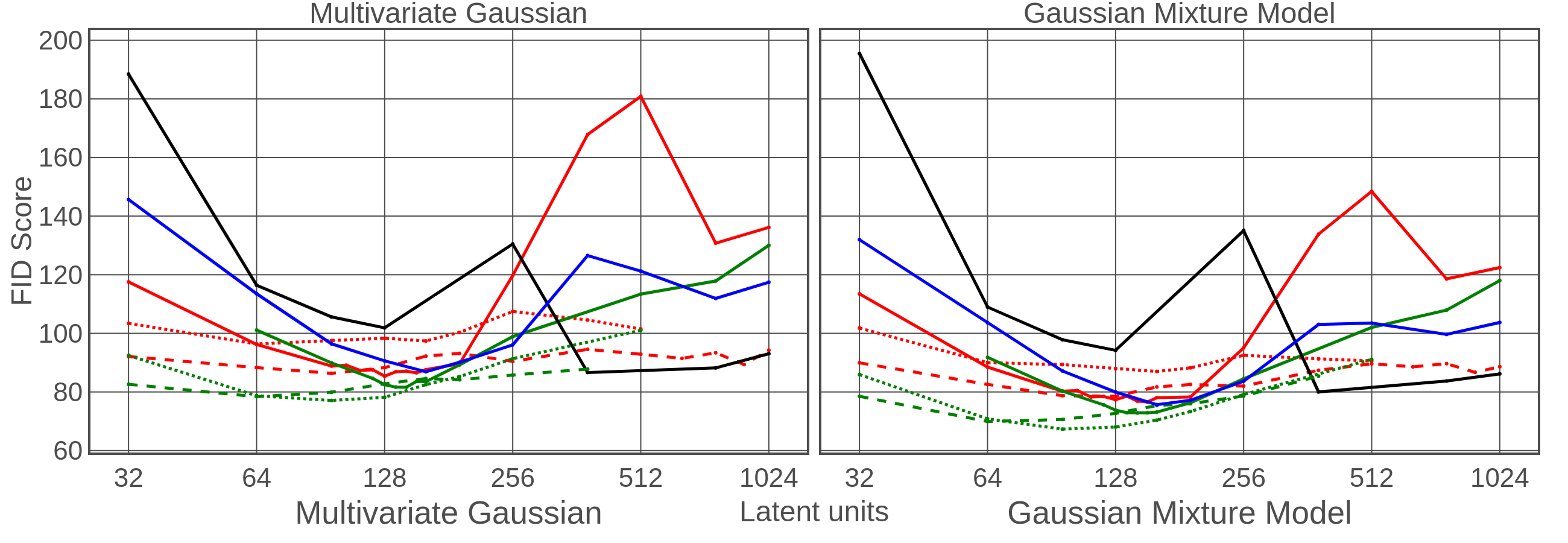}
    \caption{FID Score on CIFAR-10 for DAE with varying depth size of the latent space and a baseline VAE. VAE baseline is the solid black line,  for the DAEs filter bases are (64, red), (128, green), (256, blue), solid lines are a depth of 1, dashed lines are a depth of 2 and the dotted line is a depth of 3 for 64 filters and depths of [3,3,2,1] for 128 filters. The left graph shows FIDs for samples drawn from a MVG and a GMM on the right.}
    \label{fig:DAEdepth}
\end{figure*}
\section{Discussion}

Looking back at our hypothesis in Section \ref{sec:hypotheses} we will examine how our results support or disprove our hypotheses. In Figure \ref{fig:DAEvsVAE} we can clearly see that DAEs have the ability to outperform VAEs when we abandon the simple assumption about a VAE's prior (being a centred, isotropic, multivariate Gaussian \(p(\mathbf{z}) = \mathcal{N} (\mathbf{z; 0, I})\) with a diagonal covariance matrix) and compute the Gaussian with a full covariance matrix or use a more complex model like a GMM. DAEs show much stronger scaling performance than VAEs from 32 to 128 filters, but stall at 256 filters on CIFAR-10. This may be down to issues of training very wide networks as the biggest convolutional layer would reach 2048 filters and is supported by this model having a much lower FID score for reconstructed images from the training set, but higher FID scores for the test set images. The scaling and overall performance is better and more stable on CelebA. This is weak evidence that increasing network width produces a model that has greater smoothness and interpolatability; the improved FID scores are a consequence of this.

Deeper networks trained on CIFAR-10 also showed improvements in overall generative performance, importantly their optimum was at a smaller latent space size and their generative performance was much less sensitive to the latent space size as can be seen in Figure \ref{fig:DAEdepth}. This suggests that network depth has more of a regularising impact or encourages smoothness more strongly and so may be a more critical parameter for generating smooth and interpolatable latent spaces. This should also be taken in the context of wider work on CNNs, such as EfficientNet, that show that there is an optimal balance between width and depth of CNNs \citep{tanEfficientNetRethinkingModel2019}.

We have also predicted that larger encoders should produce more Gaussian outputs in hypothesis \ref{itm:2ndhyp}. From \citet{bubeckUniversalLawRobustness2021} we expect larger networks  to produce outputs that are more Gaussian as they increase in size and approach the lower bound for Lipschitz continuity due to isoperimetry. Due to the high dimensionality of the latent spaces, they are difficult to visualise in their entirety. However, by taking random pairs of dimensions in the latent space and plotting the encoder output on a scatter graph we can try and visualise the relationship, examples are shown in Figures \ref{fig:DAEVAELatentscattercifar10} and \ref{fig:DAEVAELatentscatterceleba} in the Appendix. All of the VAEs and DAEs trained in these experiments have distributions in the latent space that are Gaussian, however, the DAEs do not have means of 0 and standard deviations of 1, their outputs have a much wider range and vary across each latent dimension. With increasing model width there is a reduction in the gap between scores between MVG and GMM sampling, suggesting that the difference between the GMM and the MVG is reduced and so the outputs of the encoder are more Gaussian.

In the context of \citet{bubeckUniversalLawRobustness2021}'s work discussed in Section \ref{sec:widerwork}, we can estimate that for CIFAR-10 (\(d=3\times10^{3}, n=5\times10^{4}\)) and CelebA (\(d=10^{4}, n=2\times10^{5}\)) we would require models with \(10^7 - 10^8\) and \(10^7 - 10^9\) parameters respectively to meet the conditions for robustness and Lipschitz continuity, depending on the reduced dimensionality of the data. The models we have trained have encoders with \(10^6 - 5\times10^7\) and \(5\times10^6 - 3\times10^7\) parameters and decoders with \(3\times10^6 - 3\times10^7\) and \(10^7 - 8\times10^7\) parameters for CIFAR-10 and CelebA respectively. We can see that our largest encoders and decoders are approaching the lower bound defined by \citet{bubeckUniversalLawRobustness2021} for continuity and that models an order or magnitude larger would begin to clear this bound. 

This could provide an explanation for why we begin to see DAEs outperform VAEs within our experimental data; the increasing size results in an encoder that produces a smoother and more interpolatable latent space and a decoder that can map the latent space back into a smoother image output space. In this case, approaching the bounding condition for robustness results in encoders and decoders that intrinsically gain the properties of VAEs to produce continuous latent spaces that can be sampled from. The result in Figure \ref{fig:DAEfrozenlatents} provides support for this, as freezing the training of the latent layer has very little impact on the overall performance of the whole DAE, suggesting that it is the intrinsic properties of the encoder and decoder that give rise to the behaviour that we see. 

A consequence of this is that, for models approaching or passing the bounding condition for robustness and Lipschitz continuity, the variational inference process does not need to be used for producing generative models. This will make it easier to train larger autoencoders for generative modelling, as the training of DAEs is simpler and more robust than that of VAEs, and this could increase the generative performance of autoencoder models as we have shown that larger DAEs outperform VAEs for MVG and GMM sampling.

\section{Conclusion}

Recent work has show that DAEs can achieve better performance at generative modelling than VAEs, but have so far been used with novel regularisation techniques and loss functions. We have we have investigated the behaviour of classic DAEs with standard loss functions and without any other novel techniques. We have shown that classic DAEs can outperform VAEs and perform similarly to novel DAE approaches. We hypothesise that this is due to large encoder and decoder networks implicitly gaining the properties of interpolatability and Lipschitz continuity, accruing the useful properties of the VAE approach. We have shown some initial evidence that supports this conclusion. However, our work is not exhaustive in investigating the bounds of where this phenomenon occurs and so we cannot explicitly say when the VAE or DAE approach should be used. These findings do suggest that future work with generative models using the VAE approach should investigate the performance of their architectures as a DAE and that there may be significant performance gains achieved by re-implementing existing state of the art VAEs as DAEs.

\bibliography{main.bib}


\clearpage

\onecolumn

\appendix
\counterwithin{figure}{section}
\section{Appendix: latent space visualisations}


\begin{figure}[hbt!]
    \centering
    \includegraphics[trim=0 12 0 20,clip, width=0.9\textwidth]{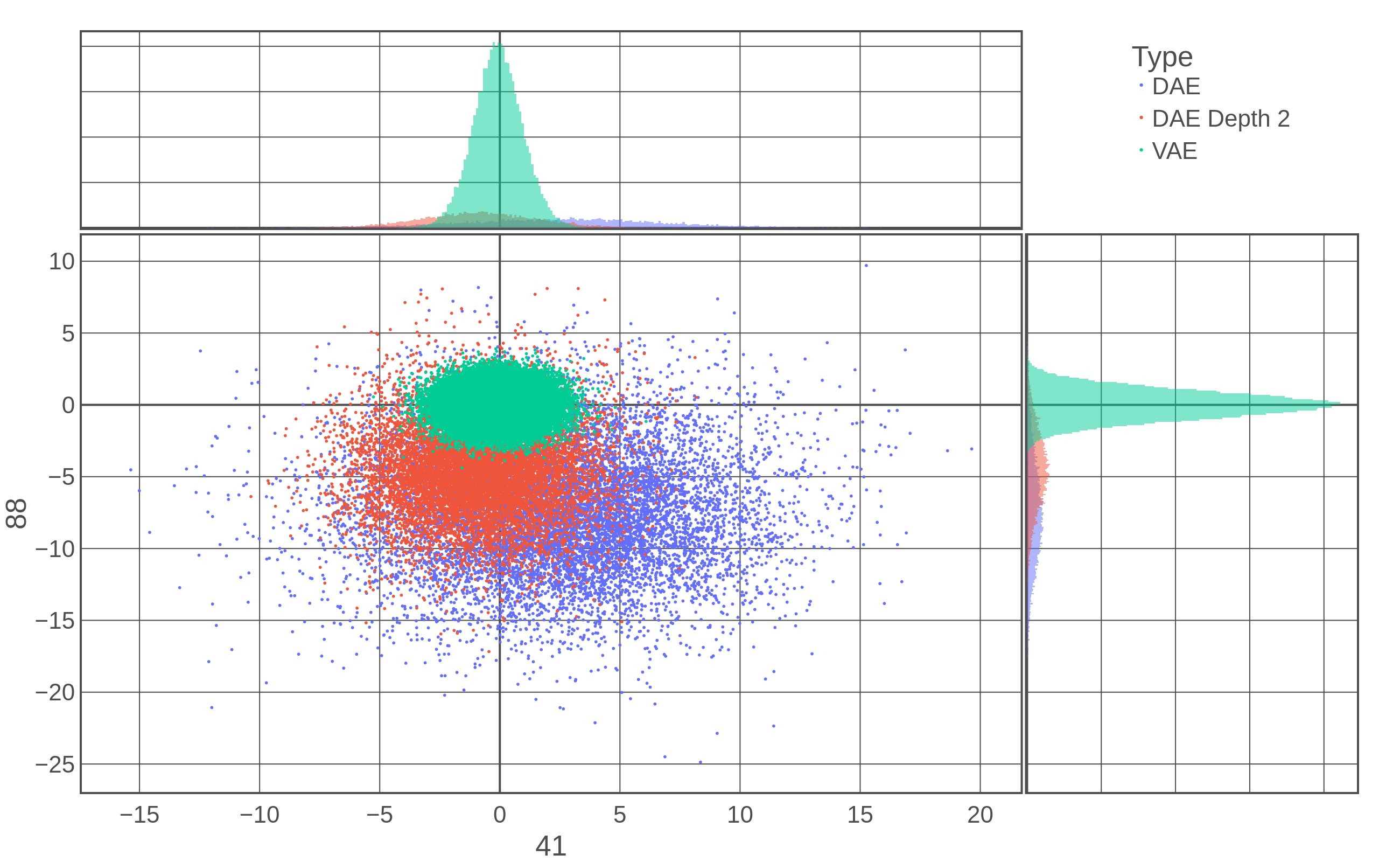}
    \caption{Scatter plot of the encoded train images for two random latent space dimensions for an a DAE (blue, red) and VAE (green) with filter base 128 and latent space dimension of 128 trained on CIFAR-10.}
    \label{fig:DAEVAELatentscattercifar10}
\end{figure}


\begin{figure}[hbt!]
    \centering
    \includegraphics[trim=0 12 0 20,clip, width=0.9\textwidth]{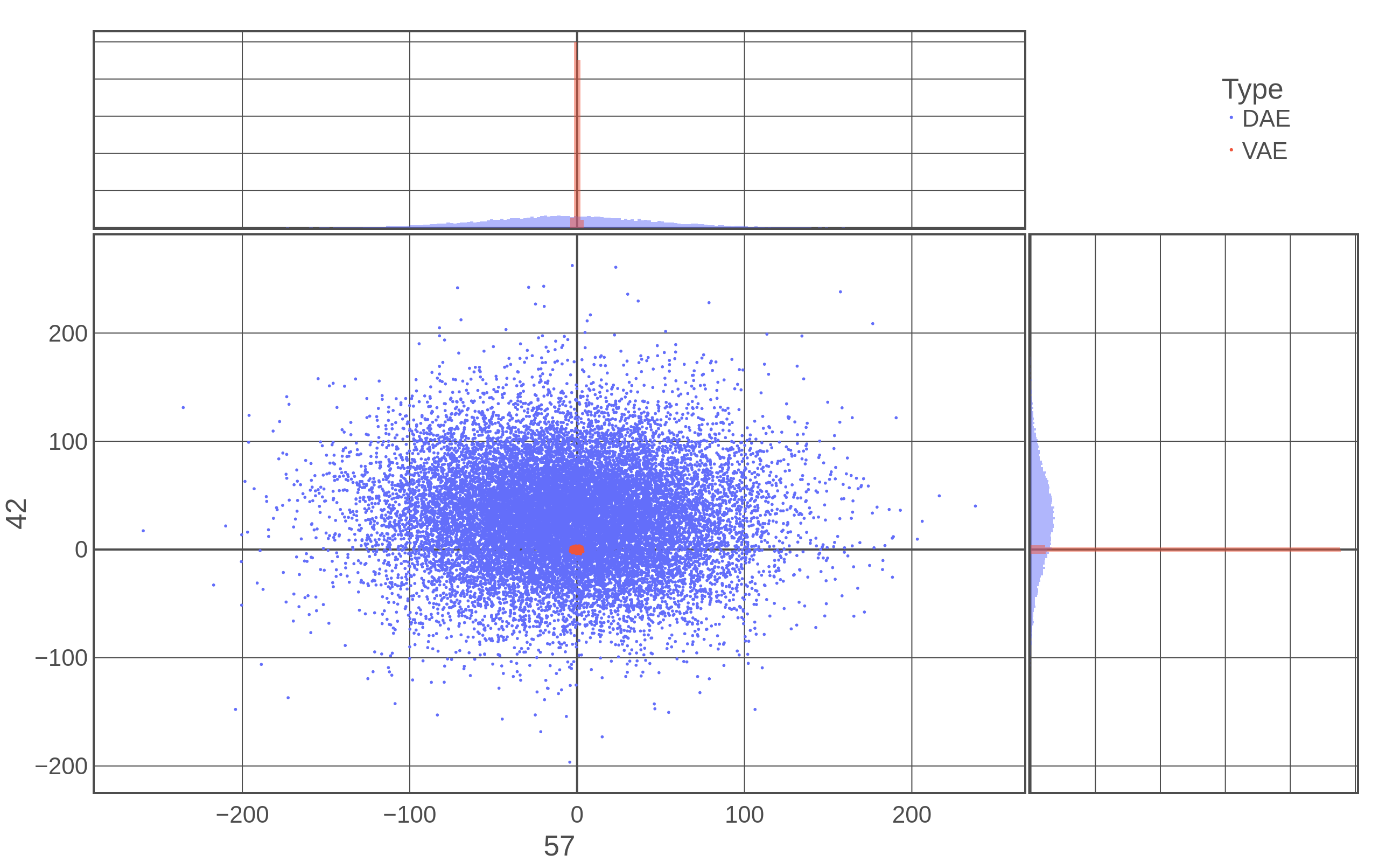}
    \caption{Scatter plot of the encoded train images for two random latent space dimensions for an a DAE (blue) and VAE (red) with filter base 128 and latent space dimension of 64 trained on CelebA.}
    \label{fig:DAEVAELatentscatterceleba}
\end{figure}

\end{document}